\documentclass{article}

\PassOptionsToPackage{numbers, sort&compress}{natbib}
\usepackage[final]{ml4mols_2020}

\usepackage[utf8]{inputenc} % allow utf-8 input
\usepackage[T1]{fontenc}    % use 8-bit T1 fonts
\usepackage{hyperref}       % hyperlinks
\usepackage{url}            % simple URL typesetting
\usepackage{booktabs}       % professional-quality tables
\usepackage{amsfonts}       % blackboard math symbols
\usepackage{nicefrac}       % compact symbols for 1/2, etc.
\usepackage{microtype}      % microtypography
\usepackage{graphicx}
\usepackage{subcaption}     % Subfigures
\usepackage{tikz}           % tikz illustrations
\usepackage{pgfplots}       % plots in tikz picture
\usepackage{pgf}            % pgf plots
\usepackage[separate-uncertainty=true]{siunitx}        % nicely formatted numbers and units
\usepackage{enumitem}       % Customize enumerate
\usepackage{amsthm}         % Proofs
\usepackage{wrapfig}        % wrapfigure
\usepackage{arydshln}       % dashed line
\usepackage{amsmath,amsfonts,bm}
\usepackage{accents} % for hattilde

\def\eqref#1{equation~\ref{#1}}
\def\1{\bm{1}}

\def\va{{\bm{a}}}
\def\vb{{\bm{b}}}

\def\ve{{\bm{e}}}

\def\vh{{\bm{h}}}

\def\vm{{\bm{m}}}

\def\vx{{\bm{x}}}

\def\vz{{\bm{z}}}

\def\mF{{\bm{F}}}

\def\mW{{\bm{W}}}

\DeclareMathAlphabet{\mathsfit}{\encodingdefault}{\sfdefault}{m}{sl}
\SetMathAlphabet{\mathsfit}{bold}{\encodingdefault}{\sfdefault}{bx}{n}
\newcommand{\tens}[1]{\bm{\mathsfit{#1}}}

\def\tW{{\tens{W}}}

\newcommand{\Cov}{\mathrm{Cov}}
\newlength{\dtildeheight}

\newlength{\dcheckheight}

\usepackage[capitalize,nameinlink]{cleveref} % better than autoref; import last

\DeclareSIUnit\debye{D}  % Debye (dipole moment)
\DeclareSIUnit\cal{cal}  % Calorie

\crefname{section}{Sec.}{Sec.}
\crefname{appendix}{App.}{App.}
\usetikzlibrary{calc,shapes,arrows.meta,angles,quotes}
\usepgfplotslibrary{fillbetween} % fill below curve
\pgfdeclarelayer{background}
\pgfdeclarelayer{foreground}
\pgfsetlayers{background,main,foreground}

\definecolor{uglyblue}{named}{blue}
\definecolor{uglygreen}{named}{green}
\definecolor{uglyred}{named}{red}
\definecolor{uglyyellow}{named}{yellow}

\definecolor{brewerpurple}{RGB}{117,112,179}
\definecolor{brewergreen}{RGB}{27,158,119}
\definecolor{brewerred}{RGB}{217,95,2}

\definecolor{blue}{RGB}{1, 115, 178}
\definecolor{orange}{RGB}{222, 143, 5}
\definecolor{green}{RGB}{2, 158, 115}
\definecolor{red}{RGB}{213, 94, 0}
\definecolor{pink}{RGB}{204, 120, 188}
\definecolor{yellow}{RGB}{236, 225, 51}

\title{Fast and Uncertainty-Aware Directional Message Passing for Non-Equilibrium Molecules}

\author{%
  Johannes Gasteiger, Shankari Giri, Johannes T. Margraf, Stephan Günnemann\\
  Technical University of Munich\\
  \texttt{\{j.gasteiger,guennemann\}@in.tum.de}, \texttt{shankari.giri@tum.de},\\ \texttt{johannes.margraf@ch.tum.de}\\
}

\begin{document}

\maketitle

\begin{abstract}
Many important tasks in chemistry revolve around molecules during reactions. This requires predictions far from the equilibrium, while most recent work in machine learning for molecules has been focused on equilibrium or near-equilibrium states. In this paper we aim to extend this scope in three ways. First, we propose the DimeNet$^{++}$ model, which is 8x faster and \SI{10}{\percent} more accurate than the original DimeNet on the QM9 benchmark of equilibrium molecules. Second, we validate DimeNet$^{++}$ on highly reactive molecules by developing the challenging \textsc{COLL} dataset, which contains distorted configurations of small molecules during collisions. Finally, we investigate ensembling and mean-variance estimation for uncertainty quantification with the goal of accelerating the exploration of the vast space of non-equilibrium structures. Our DimeNet$^{++}$ implementation as well as the \textsc{COLL} dataset are available online.\footnote{\url{https://www.daml.in.tum.de/dimenet}}
\end{abstract}

\section{Introduction} \label{sec:intro}

Modern machine learning models for molecular property prediction typically focus on molecules in equilibrium (e.g.\ QM9 \cite{ramakrishnan_quantum_2014}) or close to the equilibrium (e.g.\ MD17 \cite{chmiela_machine_2017}, ANI-1 \cite{smith_ani-1_2017}, QM7-X \cite{hoja_qm7-x_2020}). However, this precludes their application to the dynamics during chemical reactions, which involve transition states far away from the equilibrium. Making reliable predictions for these states requires models that are able to cover a much broader range of chemical and configurational space, i.e.\ including open-shell electronic structures, stretched bonds and distorted angles. In this work we aim at making progress on this problem from three directions.

First, we propose a model that is fast, accurate, and generalizes well both to different configurations and different molecules. This model predicts both the molecule's energy and the forces acting on each atom, since the latter are crucial for the molecule's dynamic behavior. To this end, we start from the recently proposed Directional Message Passing Neural Network (DimeNet) \cite{gasteiger_directional_2020}, which fulfills all of these properties except one: It is comparatively slow to compute. We perform a thorough model analysis to fix this and propose the DimeNet$^{++}$ model, which achieves an 8x runtime improvement while also improving predictions by \SI{10}{\percent} on average and by \SI{20}{\percent} for energies.

Second, we develop a new dataset that contains highly reactive non-equilibrium systems. The new \textsc{COLL} dataset contains \num{140000} configurations of pairs of molecules reacting at high kinetic energies. It only consists of small molecules but covers the space of reactions much better and includes a significantly wider range of energies and forces than previous benchmarks, as shown in \cref{fig:energies}.

Due to the vast number of possible non-equilibrium configurations it is crucial that we are able to detect when we move out of the region covered by the training data and react appropriately (e.g.\ via active learning). To achieve this we investigate ensembling \cite{hansen_neural_1990} and mean-variance estimation~\cite{nix_estimating_1994}. We conclude that both are insufficient, due to their overhead and inability of reliably predicting the energy and force uncertainties.

\section{DimeNet$^{++}$}

\begin{figure}
    \centering
    \resizebox{\textwidth}{!}{
    \input{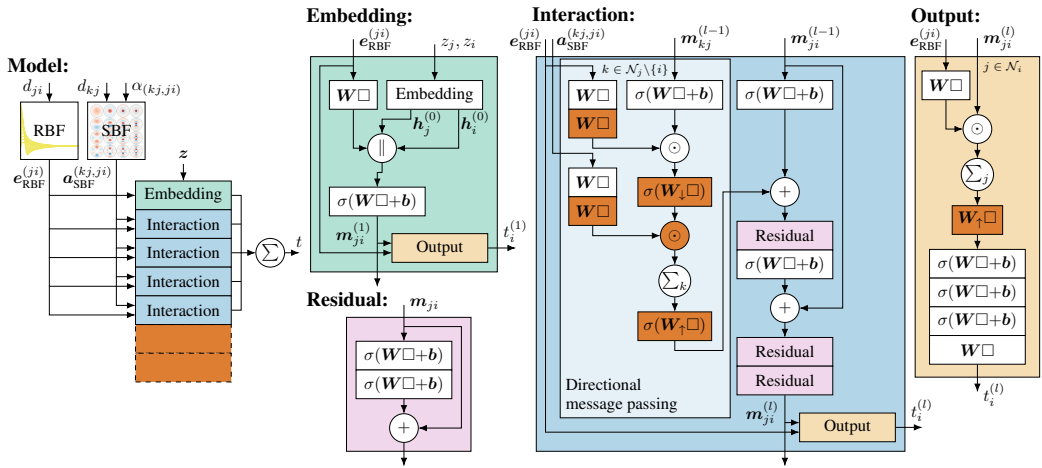}
    }
    \caption{DimeNet$^{++}$ architecture. $\square$ denotes the layer's input and $\|$ denotes concatenation. Changes to regular DimeNet are highlighted in \textbf{\textcolor{red}{red}}.}
    \label{fig:dimenet_pp}
\end{figure}

\textbf{DimeNet.} DimeNet is a recently proposed Graph Neural Network (GNN) for molecular property prediction \cite{gasteiger_directional_2020}. It improves upon regular GNNs in two ways. Normal GNNs represent each atom $i$ separately via its embedding $\vh_i$ and update these in each layer $l$ via message passing. DimeNet instead embeds and updates the messages between atoms $\vm_{ji}$, which enables it to consider directional information (via bond angles $\alpha_{(kj,ji)}$) as well as interatomic distances $d_{ji}$. DimeNet furthermore embeds distances and angles jointly using a spherical 2D Fourier-Bessel basis, resulting in the update
\begin{equation}
    \vm_{ji}^{(l+1)} = f_\text{update}(\vm_{ji}^{(l)}, \sum_{k \in \mathcal{N}_j \setminus \{i\}} f_\text{int}(\vm_{kj}^{(l)}, \ve_\text{RBF}^{(ji)}, \va_\text{SBF}^{(kj,ji)})),
    \label{eq:agg}
\end{equation}
where $f_\text{update}$ denotes the update function, $f_\text{int}$ the interaction function, $\ve_\text{RBF}^{(ji)}$ the radial basis function (RBF) representation of $d_{ji}$ and $\va_\text{SBF}^{(kj,ji)}$ the spherical basis function (SBF) representation of $d_{kj}$ and $\alpha_{(kj,ji)}$. In this work we do not touch either of those contributions and instead focus on the model architecture. The updated DimeNet$^{++}$ architecture is illustrated in \cref{fig:dimenet_pp}.

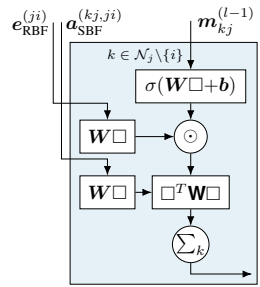
\begin{wrapfigure}[14]{r}{0.26\textwidth}
    \centering
    \vspace*{-0.3cm}
    \resizebox{0.26\textwidth}{!}{
    %!TEX root = ../dimenet_pp.tex

\pgfplotsset{set layers=standard}
\begingroup
\medmuskip=0mu
\begin{tikzpicture}
    \def\linewidth{0.4pt}  % This is just the default linewidth
    \tikzstyle{block} = [draw, line width=\linewidth, minimum height=0.6cm, minimum width=2cm, fill=white, anchor=north]
    \tikzstyle{opcircle} = [draw, line width=\linewidth, circle, fill=white, anchor=north, inner sep=0.1pt, minimum size=0.6cm]
    \tikzstyle{flow} = [-{Latex}, line width=\linewidth]
    
    % Directional message passing
    \node[block, anchor=north west, minimum width=3.4cm, minimum height=4.4cm, fill=blue!10] (int_dime_block) at (0, 0) {};
    % \node[anchor=south west, align=left] (int_dime) at (int_dime_block.south west) {Directional \\ message passing};
    
    % Input
    \node[anchor=south east, inner xsep=2pt] (int_rbf_ji) at ($(int_dime_block.north west) + (-0.3, 0)$) {$\ve_\text{RBF}^{(ji)}$};
    \node[anchor=south west, inner xsep=2pt] (int_sbf) at ($(int_dime_block.north -| int_rbf_ji.east) + (0.15, 0)$) {$\va_\text{SBF}^{(kj,ji)}$};
    \node[anchor=south west] (int_mkj) at ($(int_dime_block.north -| int_sbf.west) + (2.35, 0)$) {$\vm_{kj}^{(l-1)}$};
    \node[anchor=north east] (int_Nj) at (int_mkj.west |- int_dime_block.north) {\scriptsize $k \in \mathcal{N}_j \setminus \{i\}$};

    % Dense layers messages
    \node[block] (int_dense_mkj) at (int_mkj.south west |- int_dense_mji.north) {$\sigma(\mW \square + \vb)$};
    \draw[flow] (int_mkj.west) -> (int_dense_mkj);

    % RBF
    \node[opcircle] (int_dot_rbf) at ($(int_dense_mkj.south) + (0, -0.3)$) {$\odot$};
    \node[block, anchor=north east, minimum width=1cm] (int_dense_rbf) at ($(int_dense_mkj.west |- int_dot_rbf.north)$) {$\mW \square$};
    \draw[flow] (int_rbf_ji.east) |- ($(int_dense_rbf.north) + (0, 0.3)$) -> (int_dense_rbf.north);
    \draw[flow] (int_dense_mkj) -> (int_dot_rbf);
    \draw[flow] (int_dense_rbf) |- (int_dot_rbf.west);

    % SBF
    \node[block, minimum width=1.4cm] (int_bilinear_sbf) at ($(int_dot_rbf.south) + (0, -0.4)$) {$\square^T \tW \square$};
    \node[block, minimum width=1cm] (int_dense_sbf) at (int_dense_rbf |- int_bilinear_sbf.north) {$\mW \square$};
    \draw[flow] (int_sbf.west) |- ($(int_dense_sbf.north) + (0, 0.3)$) -> (int_dense_sbf);
    \draw[flow] (int_dot_rbf.south) -> (int_bilinear_sbf);
    \draw[flow] (int_dense_sbf) -> (int_bilinear_sbf.west);

    % Sum up m_kj
    \node[opcircle, anchor=north] (int_sum_mkj) at ($(int_bilinear_sbf.south) + (0, -0.3)$)  {$\sum_k$};
    \draw[flow] (int_bilinear_sbf) -> (int_sum_mkj);
    \draw[flow] (int_sum_mkj) |- ($(int_dime_block.east |- int_sum_mkj.south) + (-0.1, -0.2)$);
\end{tikzpicture}
\endgroup
    }
    \caption{DimeNet's original ``directional message passing'' block.}
    \label{fig:dimenet}
\end{wrapfigure}
\textbf{Combinatorial representation explosion.} DimeNet embeds every message, i.e.\ every interacting \emph{pair}, separately and thus uses many times as many embeddings as a regular GNN. This combinatorial explosion becomes even worse in the interaction block, where we need to embed every \emph{triplet} to represent bond angles. On the QM9 dataset (with \SI{5}{\angstrom} cutoff) we found that DimeNet uses around 15x as many message embeddings as there are atoms and again around 15x as many triplet representations. Operations in the ``directional message passing'' block are thus 15x more expensive than elsewhere in the model, while those in the output block (which uses atom embeddings) are 15x cheaper.

\textbf{Fast interactions.} We therefore first focus on the expensive ``directional message passing'' block. It is DimeNet's centerpiece, modelling the interaction between embeddings $\vm_{kj}$ and basis representations $\ve_\text{RBF}^{(ji)}$ and $\va_\text{SBF}^{(kj,ji)}$. As such, it requires an adequately expressive transformation. The original DimeNet accomplishes this with a bilinear layer, as shown in \cref{fig:dimenet}. Unfortunately, this layer is very expensive, which is exacerbated by being used in the model's most costly component. We alleviate this by replacing it with a simple Hadamard product and compensate for the loss in expressiveness by adding multilayer perceptrons (MLPs) for the basis representations. This recovers the original accuracy at a fraction of the computational cost (see \cref{sec:exp}).

\textbf{Embedding hierarchy.} We can directly leverage the fact that certain parts of the model use a higher number of embeddings by reducing the embedding size in these parts via down- and upprojection layers $\mW_{\downarrow}$ and $\mW_{\uparrow}$. This both accelerates the model and removes information bottlenecks, since we no longer aggregate information to a smaller number of equally sized embeddings.

\textbf{Other improvements.} We furthermore found that using 4 layers performs en par with the original 6 for $U_0$. Moreover, larger batch sizes significantly slowed down convergence, and mixed precision caused the model's precision to break down completely. Considering that DimeNet's relative error is below float16's machine precision ($5 \cdot 10^{-4}$), the latter might be expected.

\section{\textsc{COLL} Dataset}

\begin{figure}
    \centering
    \begin{minipage}[t]{0.65\textwidth}
        \includegraphics[width=\textwidth,keepaspectratio]{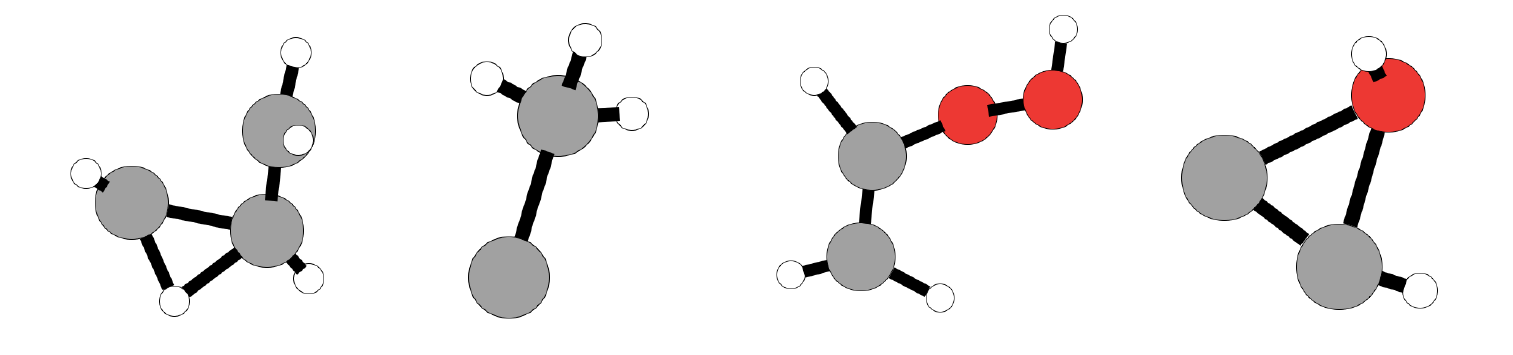}
       \caption{Example configurations from the \textsc{COLL} dataset. \textsc{COLL} covers a much broader range of configurational space, including stretched bonds and distorted angles.}
       \label{fig:example_molecules}
    \end{minipage}
    \hfill
    \begin{minipage}[t]{0.325\textwidth}
        \centering
        \input{figures/coll_atomization_energy_v1.2.pgf}
        \caption{Distribution of atomization energy per atom. \textsc{COLL} covers a much wider range.}
        \label{fig:energies}
    \end{minipage}
\end{figure}

The \textsc{COLL} dataset consists of configurations taken from molecular dynamics simulations of molecular collisions. To this end, collision simulations were performed with the cost-effective semiempirical GFN2-xTB method \cite{bannwarth_gfn2-xtbaccurate_2019}. Subsequently, energies and forces for \num{140000} random snapshots taken from these trajectories were recomputed with density functional theory (DFT). These calculations were performed with the revPBE functional and def2-TZVP basis, including D3 dispersion corrections~\cite{zhang_comment_1998}.

Exemplary structures from the \textsc{COLL} set are shown in \cref{fig:example_molecules}. Unlike established molecular benchmark sets (e.g.\ QM9), which consist of equilibrium or near-equilibrium configurations, the structures in \textsc{COLL} can be highly distorted. In particular, stretched bonds and angles, as well as open-shell electronic structures are prevalent. All calculations are preformed with broken spin-symmetry.

Overall, the configurations in \textsc{COLL} represent a challenge for electronic structure calculations, since such systems may display multiple self-consistent-field (SCF) solutions. This can pose a significant problem for ML-based models, as the corresponding reference potential energy surfaces can be discontinuous. To avoid this issue, multiple calculations from randomized initial wavefunctions were conducted, and the lowest energy solution selected. Furthermore, Fermi-smearing with an electronic temperature of \SI{5000}{\kelvin} was applied, which is helpful both for SCF convergence and the approximate description of static correlation effects \cite{grimme_practicable_2015}.

\section{Uncertainty Quantification}

The vast number of non-equilibrium states reachable in high-energy molecular dynamics simulations (such as reactions) means that systems will often move outside the space covered by our training set. We therefore need a reliable way of detecting a degradation in predictive performance. Most uncertainty quantification (UQ) methods are focused on providing an uncertainty estimate for the direct prediction \cite{musil_fast_2019,hirschfeld_uncertainty_2020}. However, out-of-equilibrium dynamics require uncertainty estimates for both the energy $E$ and the force $\mF = -\frac{\partial E}{\partial \vx}$. Many non-differentiable methods (e.g.\ combining GNNs with a random forest) are therefore not applicable. Ensembling is a notable exception but introduces a large computational overhead since we need to calculate predictions using multiple separate models.

Even if the method is differentiable it might only provide a mean and standard deviation, i.e.\ $\mu_E$ and $\sigma_E$ (e.g.\ mean-variance estimation (MVE)). This allows us to obtain the force prediction via
\begin{equation}
    \mu_\mF = \left< -\frac{\partial E}{\partial \vx} \right> = -\frac{\partial}{\partial \vx}\left< E \right> = -\frac{\partial}{\partial \vx} \mu_E.
\end{equation}
However, performing the same operation on $\sigma_E$ does not yield the analogous result:
\begin{equation}
    \frac{\partial}{\partial \vx} \sigma_E^2 = \frac{\partial}{\partial \vx} \left( \left<E^2\right> - \left<E\right>^2 \right) = -2 \left( \left< -E \frac{\partial}{\partial \vx} E \right> - \left<E\right> \left< -\frac{\partial}{\partial \vx} E \right> \right) = -2 \Cov(E, \mF).
\end{equation}
There is thus no general way of estimating $\sigma_\mF$ for these kinds of models. Instead, we have to rely on $\sigma_E$ as the uncertainty measure and hope that it correlates with the force error.

%However, we can make use of the fact that we do have easy access to the covariance $\Cov(E, \mF)$. By including this term in the loss function we can enforce a high correlation between force and energy in model space. This allows us to estimate the force error \emph{directly} via $\sigma_E$. We thus propose to minimize the loss function
%\begin{equation}
%    \mathcal{L} = \text{NLL}_E + \rho \text{MAE}_\mF - \gamma \Cov(E, \mF) = \frac{1}{2} \ln{\sigma_E^2} + \frac{(E_t - \mu_E)^2}{2\sigma_E^2} + \rho \left| \mF_t - \mu_\mF \right| + \frac{\gamma}{2} \frac{\partial}{\partial \vx} \sigma_E^2
%\end{equation}
%for training and monitor $\sigma_E$ to detect anomalies, which we found to be suprisingly effective in practice.

\section{Experiments} \label{sec:exp}

\begin{table}
    \centering
    \caption{$U_0$ validation MAE on the QM9 dataset after each DimeNet$^{++}$ improvement.}
    \begin{tabular}{l@{\hspace{0.2cm}}c@{\hspace{0.2cm}}c@{\hspace{0.2cm}}}
	{Model} & {Time per epoch / min} & {Val MAE $U_0$ / meV}\\
	\hline
	DimeNet & 35.4 & 8.27 \\
	\& Hadamard product & 6.6 & 9.45 \\
	\& 2-layer MLP for RBF and SBF & 7.1 & 8.42 \\
	\& reduced to 4 layers & 4.7 & 8.54 \\
	\& triplet embedding size to 64 & 4.2 & 7.60 \\
	\& output (atom) embedding size to 256 & 4.3 & 7.05 \\
\end{tabular}
% Time per 10k steps converted via (110_000 / 32 + 10_000 / 32 / 2) / (10_000 + 10_000 / 32 / 2) = 0.353846
    \label{tab:dimenet_pp_improvements}
\end{table}

\textbf{DimeNet$^{++}$} In \cref{tab:dimenet_pp_improvements} we evaluate each of the proposed DimeNet improvements separately on the $U_0$ validation set of QM9 \cite{ramakrishnan_quantum_2014}. We see that each change either reduces the runtime or improves the error. Exchanging the bilinear layer for a Hadamard product has by far the largest impact, single-handedly decreasing the runtime by a factor of 5. Interestingly, decreasing the embedding size both accelerates the model and improves the accuracy. This is either due to the additional down- and upprojection layers improving expressiveness or to the smaller embedding size improving generalization.

\begin{table}
    \centering
    \caption{MAE on the QM9 dataset. DimeNet$^{++}$ performs best on average and for most properties individually, despite being 8x faster than DimeNet.}
    \begin{tabular}{lcccccc}
Target &                                             Unit &                 SchNet &                   MGCN &            DeepMoleNet &       DimeNet & \textbf{DimeNet$^{++}$} \\
\hline
$\mu$                        &                                      \si{\debye} &           \num{0.0330} &           \num{0.0560} &  \textbf{\num{0.0253}} &  \num{0.0286} &            \num{0.0297} \\
$\alpha$                     &                                     \si{\bohr^3} &            \num{0.235} &  \textbf{\num{0.0300}} &           \num{0.0681} &  \num{0.0469} &            \num{0.0435} \\
$\epsilon_\text{HOMO}$       &                         \si{\milli\electronvolt} &             \num{41.0} &             \num{42.1} &    \textbf{\num{23.9}} &    \num{27.8} &              \num{24.6} \\
$\epsilon_\text{LUMO}$       &                         \si{\milli\electronvolt} &             \num{34.0} &             \num{57.4} &             \num{22.7} &    \num{19.7} &     \textbf{\num{19.5}} \\
$\Delta\epsilon$             &                         \si{\milli\electronvolt} &             \num{63.0} &             \num{64.2} &             \num{33.2} &    \num{34.8} &     \textbf{\num{32.6}} \\
$\left< R^2 \right>$         &                                     \si{\bohr^2} &  \textbf{\num{0.073}} &            \num{0.110} &            \num{0.680} &   \num{0.331} &             \num{0.331} \\
ZPVE                         &                         \si{\milli\electronvolt} &             \num{1.70} &    \textbf{\num{1.12}} &             \num{1.90} &    \num{1.29} &              \num{1.21} \\
$U_0$                        &                         \si{\milli\electronvolt} &             \num{14.0} &             \num{12.9} &             \num{7.70} &    \num{8.02} &     \textbf{\num{6.32}} \\
$U$                          &                         \si{\milli\electronvolt} &             \num{19.0} &             \num{14.4} &             \num{7.80} &    \num{7.89} &     \textbf{\num{6.28}} \\
$H$                          &                         \si{\milli\electronvolt} &             \num{14.0} &             \num{14.6} &             \num{7.80} &    \num{8.11} &     \textbf{\num{6.53}} \\
$G$                          &                         \si{\milli\electronvolt} &             \num{14.0} &             \num{16.2} &             \num{8.60} &    \num{8.98} &     \textbf{\num{7.56}} \\
$c_\text{v}$ \vspace{1pt}    & \si[per-mode=fraction]{\cal\per\mol\per\kelvin} &           \num{0.0330} &           \num{0.0380} &           \num{0.0290} &  \num{0.0249} &   \textbf{\num{0.0230}} \\
\hline
std. MAE \rule{0pt}{0.9em} &                                    \si{\percent} &             \num{1.76} &             \num{1.86} &             \num{1.03} &    \num{1.05} &     \textbf{\num{0.98}} \\
logMAE                       &                                                - &            \num{-5.17} &            \num{-5.26} &            \num{-5.46} &   \num{-5.57} &    \textbf{\num{-5.67}} \\
\end{tabular}
    \label{tab:dimenet_pp}
\end{table}

We evaluate the final DimeNet$^{++}$ model on all QM9 targets and compare it to the state-of-the-art models SchNet \cite{schutt_schnet:_2017}, MGCN \cite{lu_molecular_2019}, and DeepMoleNet \cite{liu_transferable_2020}. \cref{tab:dimenet_pp} shows that DimeNet$^{++}$ performs better for most targets and best overall, in addition to being 8x faster than DimeNet. OrbNet \cite{qiao_orbnet_2020} performs better on $U_0$, but has not published results for the other properties. Note that the DFT-based representations introduced by DeepMoleNet and OrbNet can also be incorporated into DimeNet$^{++}$.

\begin{table}
    \centering
    \caption{Performance on the \textsc{COLL} dataset. MAE is given in \si{\electronvolt} and \si[per-mode=symbol]{\electronvolt\per\angstrom}. $\rho$ denotes the correlation coefficient and $\Delta$ the absolute error. DimeNet$^{++}$ performs significantly better than SchNet. MVE is much faster than ensembling but unable to estimate the force error.}
    \begin{tabular}{l@{\hspace{0.05cm}}c@{\hspace{0.2cm}}c@{\hspace{0.2cm}}c@{\hspace{0.2cm}}c@{\hspace{0.1cm}}c@{\hspace{0.1cm}}c@{\hspace{0.1cm}}}
	 & Time per epoch & MAE$_E$ & MAE$_\mF$ & {$\rho(\Delta_E,\sigma_E)$} & {$\rho(\Delta_\mF,\sigma_\mF)$} & {$\rho(\Delta_\mF,\sigma_E)$}\\
	\hline
	SchNet & \textbf{\SI{8.9}{\minute}} & 0.198 & 0.172 & - & - & -\\
	DimeNet$^{++}$ & \SI{10.4}{\minute} & 0.047 & 0.040 & - & - & -\\
	Ensemble (DimeNet$^{++}$) & \SI{31.2}{\minute} & 0.050 & \textbf{0.038} &  0.42  &  0.85 & 0.64\\
	MVE (DimeNet$^{++}$) & \SI{13.6}{\minute} & \textbf{0.033} & 0.041 & 0.16 & - & 0.05\\
\end{tabular}
% Time per 1k steps converted via (120_000 / 32 + 10_000 / 32 / 2) / (1_000 + 10_000 / 32 / 2) = 3.3783783783783785
% Time per 10k steps converted via (120_000 / 32 + 10_000 / 32 / 2) / (10_000 + 10_000 / 32 / 2) = 0.38461538461538464
% Scale to 70k+10k: (70_000 + 10_000 / 2) / (9_000 + 1_000 / 2)

% SchNet MAE: val: energy=0.1928, force=0.1697; test: energy=0.1979, force=0.1723
% SchNet RMSE: val: energy=0.2761, force=0.2752; test: energy=0.2867, force=0.2807

    \label{tab:uq}
\end{table}

\textbf{\textsc{COLL} dataset.} \cref{tab:uq} shows that DimeNet$^{++}$ is only \SI{17}{\percent} slower than SchNet (reference implementation), while reducing the error by \SI{76}{\percent} on average. As expected, the \textsc{COLL} dataset is significantly more challenging than QM9. Both SchNet and DimeNet$^{++}$ exhibit an MAE that is around 10x higher than on QM9.

\textbf{Uncertainty quantification.} Ensembling and MVE both struggle with estimating the energy uncertainty, as shown for DimeNet$^{++}$ in \cref{tab:uq}. The force error is very well estimated by the ensemble, but not by the energy uncertainty -- especially for MVE. The energy uncertainty is thus not as good a proxy for the force error as one would expect. While the ensemble does perform decently, its computational overhead is still considerable. Reliable and fast uncertainty estimates thus remain an important direction for future work.

% \section{Conclusion}

% We proposed the DimeNet$^{++}$ model, a fast and accurate model that performs well on out-of-equilibrium predictions. To test this capacity we generated the challenging \textsc{COLL} dataset, which is based on molecular collisions far away from the equilibrium. We found both ensembling and MVE to be insufficient for estimating uncertainty in this setting, demonstrating that reliable uncertainty estimates are an important direction for future work.

\begin{ack}
This research was supported by the TUM International Graduate School of Science and Engineering (IGSSE), GSC 81. The authors of this work take full responsibilities for its content.
\end{ack}

%\small
\bibliography{dimenet_pp}
\bibliographystyle{dimenet_pp}
\normalsize
\end{document}